%% file: iclr2026_conference.tex
\documentclass{article} 
\usepackage{iclr2026_conference, times}
\iclrfinalcopy
\input{math_commands.tex}

\usepackage{hyperref}
\usepackage{url}
\usepackage{graphicx}
\usepackage{subcaption}
\usepackage{booktabs}
\title{Towards Efficient and Stable Ocean State Forecasting: A Continuous-Time Koopman Approach}

\author{Rares Grozavescu \\
Department of Engineering\\
University of Cambridge\\
Cambridge, UK \\
\texttt{\{rg625\}@cam.ac.uk} \\
\And
Pengyu Zhang \\
Department of Engineering\\
University of Cambridge\\
Cambridge, UK \\
\texttt{\{pz281\}@cam.ac.uk} \\
\And
Mark Girolami\\
Department of Engineering\\
University of Cambridge\\
Cambridge, UK; \\
Alan Turing Institute\\
London, UK\\
\And
Etienne Meunier \\
INRIA\\
Paris, France\\
}

\begin{document}

\maketitle


\begin{abstract}
We investigate the Continuous-Time Koopman Autoencoder (CT-KAE) as a lightweight surrogate model for long-horizon ocean state forecasting in a two-layer quasi-geostrophic (QG) system. By projecting nonlinear dynamics into a latent space governed by a linear ordinary differential equation, the model enforces structured and interpretable temporal evolution while enabling temporally resolution-invariant forecasting via a matrix exponential formulation. Across 2083-day rollouts, CT-KAE exhibits bounded error growth and stable large-scale statistics, in contrast to autoregressive Transformer baselines which exhibit gradual error amplification and energy drift over long rollouts. While fine-scale turbulent structures are partially dissipated, bulk energy spectra, enstrophy evolution, and autocorrelation structure remain consistent over long horizons. The model achieves orders-of-magnitude faster inference compared to the numerical solver, suggesting that continuous-time Koopman surrogates offer a promising backbone for efficient and stable physical–machine learning climate models.
\end{abstract}

\section{Introduction}




Ocean forecasting underpins understanding of Earth's climate, yet high-fidelity Global Climate Models (GCMs) \citep{marshall1997finite,madec2015nemo} are computationally expensive, limiting large ensemble simulations. Recent data-driven surrogates based on large autoregressive architectures \citep{pathak2022fourcastnet,lam2023learning,bi2023accurate,rajabi2025sea} have shown strong short-term predictive skill but often suffer from instability over long rollout horizons.

We explore a complementary physics-structured alternative based on \textbf{Koopman Theory} \citep{koopman1931}. Specifically, we apply the \textbf{Continuous-Time Koopman Autoencoder (CT-KAE)} \citep{grozavescu2026continuous} to a two-layer quasi-geostrophic (QG) model. By evolving the latent representation via a linear Ordinary Differential Equation (ODE), the model enforces explicit operator structure while enabling temporally resolution-invariant forecasting through a matrix exponential formulation.

Rather than maximizing short-term pixel-wise accuracy, we prioritize long-horizon dynamical consistency: bounded error growth and preservation of bulk statistical invariants. The central question we address is whether explicit linear structure in latent space improves stability without sacrificing computational efficiency.

\section{The Quasi-Geostrophic Model}

The Quasi-Geostrophic (QG) model have been used extensively in oceanography for modelling midlatitude oceanic circulation \citep{medjo2000numerical,majda2006nonlinear}. 

\subsection{Governing Equations and Physics}

We utilize a two-layer version of the QG model where the prognostic variable is potential vorticity, $q_m$, for layers $m \in \{1,2\}$. The governing equations relate $q_m$ to the streamfunction $\psi_m$:
\begin{equation}
\label{eq:qg_equations}
q_1 = \nabla^2 \psi_1 + F_1 (\psi_2 - \psi_1), \quad \quad q_2 = \nabla^2 \psi_2 + F_2 (\psi_1 - \psi_2),
\end{equation}
where $F_1 = k_d^2 / (1+\delta)$, $F_2 = \delta F_1$, and $k_d^2 = (f_0^2/g')(H_1+H_2)/(H_1 H_2)$. Here, $H_m$ is the layer depth, and $f_0$ is the Coriolis frequency. The fluid velocity is related to $\psi_m$ via $\mathbf{u}_m = (u_m,v_m)=(-\partial_y\psi_m, \partial_x\psi_m)$. The temporal evolution of the system is governed by:
\begin{equation}
\label{eq:qg_equations2}
\partial_t q_m + \mathbf{J}(\psi_m, q_m) + \beta \partial_x\psi_m + \partial_yQ_m \partial_x\psi_m + U_m \partial_xq_m = \mathcal{D}_m
\end{equation}
where $\mathbf{J}(A,B) = \partial_xA\partial_yB - \partial_yA\partial_xB$ is the horizontal Jacobian, $Q_m$ is background potential vorticity, and $U_m$ is the background zonal mean flow velocity. The term $\mathcal{D}_m$ represents small-scale dissipation, with the lower layer additionally incorporating Ekman bottom drag such that $\mathcal{D}_2 = -r_{ek}\nabla^2 \psi_2 + \text{ssd}$.

\subsection{Related Work on QG and Machine Learning}

Machine learning approaches to QG dynamics broadly fall into two categories: subgrid-scale (SGS) parameterization and full state forecasting.

\textbf{SGS Parameterization.} Prior work has focused on learning closure terms representing unresolved processes. CNN-based approaches \citep{bolton2019applications} and stability-aware refinements \citep{frezat2022posteriori,ross2023benchmarking} have demonstrated promising results for coarse-resolution modeling.

\textbf{Full State Forecasting.} Deep learning models including CNNs, GANs, Transformers, and reduced-order hybrids have been applied to turbulent flow prediction \citep{deo2023cnn,lee2019deep,hemmasian2023transformer,chattopadhyay2020deep,besabe2025data}. More recently, diffusion-based approaches such as the Thermalizer \citep{pedersen2025thermalizer} have targeted long-horizon stability.

In contrast, our approach enforces explicit linear structure and conduct temporal evolution in latent space, promoting stability while reducing inference cost to a matrix–vector multiplication.

\section{Methodology}
\begin{figure*}
    \centering
    \includegraphics[width=0.8\linewidth]{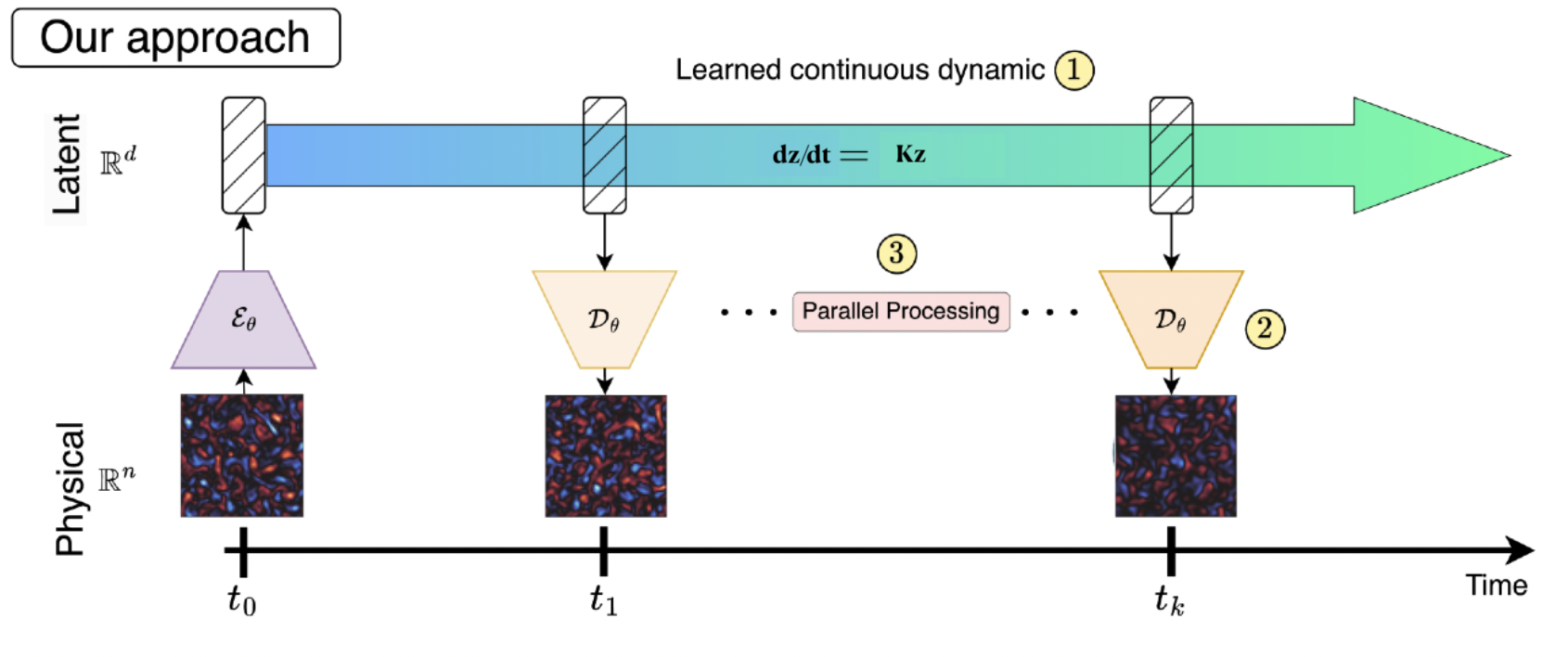}
    \caption{\textbf{CT-KAE architecture.} High-dimensional QG states $x_t$ are encoded into a latent representation $z_t$ using a dual-stream encoder that incorporates both present and historical information. The latent state evolves according to a linear continuous-time ODE $\dot{z} = \mathbf{K}z$, enabling resolution-invariant temporal querying via the matrix exponential. The decoder reconstructs physical fields from latent space.} 
    \label{fig:model}
\end{figure*}

The CT-KAE projects high-dimensional QG states into a latent space governed by a linear continuous-time dynamical system before decoding back to the physical domain. The architecture consists of a dual-stream encoder, a latent ODE module, and a decoder (Figure~\ref{fig:model}).

\textbf{Encoder and Decoder.} Let $x_t \in \mathbb{R}^{H \times W \times C}$ denote the ocean state at time $t$, comprising channels for streamfunction $\psi$ and potential vorticity $q$. CNN-based encoders are used, with a present encoder $\mathcal{E}_{\text{pres}}$ and a history encoder $\mathcal{E}_{\text{hist}}$ which take in $x_t$ and $x_{t-1}$ respectively, producing latent variable $z_t = \frac{1}{2}(\mathcal{E_{\text{pres}}}(x_t)+\mathcal{E_{\text{hist}}}(x_{t-1}))$. The decoder $\mathcal{D}$ reconstructs the physical state, $\hat{x}_t = \mathcal{D}(z_t)$.

\textbf{Continuous-Time Latent Dynamics.} In contrast to discrete-time KAEs \citep{lusch2018koopman}, which propagate the latent state via a fixed-step mapping ($z_{t+1} = \mathbf{K}z_t$), the CT-KAE governs the latent evolution through a linear ODE:
\begin{equation}
    \frac{dz}{dt} = \mathbf{K} z,
\end{equation}
where a fourth-order Runge-Kutta (RK4) scheme is employed for numerical integration. The state at any future time $t + \tau$ can also be computed via the matrix exponential:
\begin{equation}\label{eq:matrix_exp}
    z(t+\tau) = \exp(\mathbf{K} \tau) z(t).
\end{equation}
This approach allows the model to be queried at arbitrary time. We empirically evaluate this resolution-invariant capability in Section \ref{section:results}.

\textbf{Baseline: ViT (AR).}
We compare against a pure autoregressive Vision Transformer with 12 encoder layers (128-dim embeddings, 8 heads) operating on $8\times8$ patches. The model performs stepwise rollouts by recursively feeding predicted frames as input. Training uses AdamW (LR $2\times10^{-4}$, batch size 32) with the same data split as CT-KAE.

While effective at short-term prediction, the autoregressive nature of ViT leads to compounding errors over long horizons, as illustrated in Section~\ref{section:results}. This provides a strong baseline for assessing the long-term stability benefits of the CT-KAE's continuous-time latent dynamics.

\textbf{Training.} 
During training, models are optimized using short rollouts of 10 steps. Long-horizon behavior is therefore entirely emergent and not explicitly optimized during training. This setup allows us to evaluate whether stability arises from architectural structure rather than training heuristics.




\section{Dataset}

We generate data from a two-layer QG model following \citep{pedersen2025thermalizer}. The flow is driven by a background zonal mean flow ($U_1 = 0.025$ m/s, $U_2 = 0$ m/s) with physical parameters $\beta = 1.5 \times 10^{-11} \text{ m}^{-1}\text{s}^{-1}$ and linear bottom drag $r_{ek} = 5.787 \times 10^{-7} \text{ s}^{-1}$. Ground truth simulations are generated with a $3600$s integration step , then downsampled to $64 \times 64$ and subsampled every 5 steps for an effective resolution $\Delta t = 5h$. The system is integrated for 40,000 days to reach a stationary turbulent regime. We train on short 10-step trajectory segments and evaluate on unseen initial conditions using 2083-day ($\approx 10,000$ steps) rollouts without reinitialization. All fields are normalized by the training-set mean and standard deviation prior to model ingestion.

\section{Results \label{section:results}}

Because the QG system is chaotic, deterministic trajectory alignment becomes ill-posed at multi-year lead times. We therefore evaluate CT-KAE along two axes: (i) forecasting fidelity, from short-term accuracy to long-horizon invariants, and (ii) operational advantages, namely computational efficiency and temporal resolution invariance.

\subsection{Long-Horizon Emulation and Physical Invariants}

We compare long-term reconstruction accuracy using RMSE (Table~\ref{tab:metrics}). CT-KAE achieves slightly lower error than the autoregressive ViT baseline (0.938 $\pm$ 0.031 vs. 0.993 $\pm$ 0.033), indicating that latent linear structure does not compromise predictive fidelity. At 2083-day lead times, anomaly correlation (ACC) is near zero for both models (Table~\ref{tab:metrics}). This is expected: the decorrelation time scale of the QG system ($\mathcal{O}(10)$ days) is vastly exceeded, rendering deterministic trajectory alignment physically meaningless. Under such conditions, fidelity must be assessed through stability and preservation of statistical invariants rather than pointwise agreement.

To quantify exponential divergence, we define the long-horizon error growth rate:
\begin{equation}
\lambda = \frac{1}{T} \log \frac{\|x_T - \hat{x}_T\|}{\|x_0 - \hat{x}_0\|}.
\end{equation}
A positive $\lambda$ indicates exponential error amplification under recursive rollout, while a negative $\lambda$ implies contraction toward a bounded invariant set. The autoregressive ViT exhibits positive growth ($\lambda = 0.0217$), consistent with gradual exponential error amplification under recursive rollout. In contrast, CT-KAE yields $\lambda = -0.0267$, demonstrating bounded long-horizon behavior. Notably, both models are trained only on 10-step segments; the stability of CT-KAE emerges from its continuous-time operator structure rather than explicit long-horizon optimization.

\begin{figure*}[ht!]
    \centering
    \includegraphics[width=\linewidth]{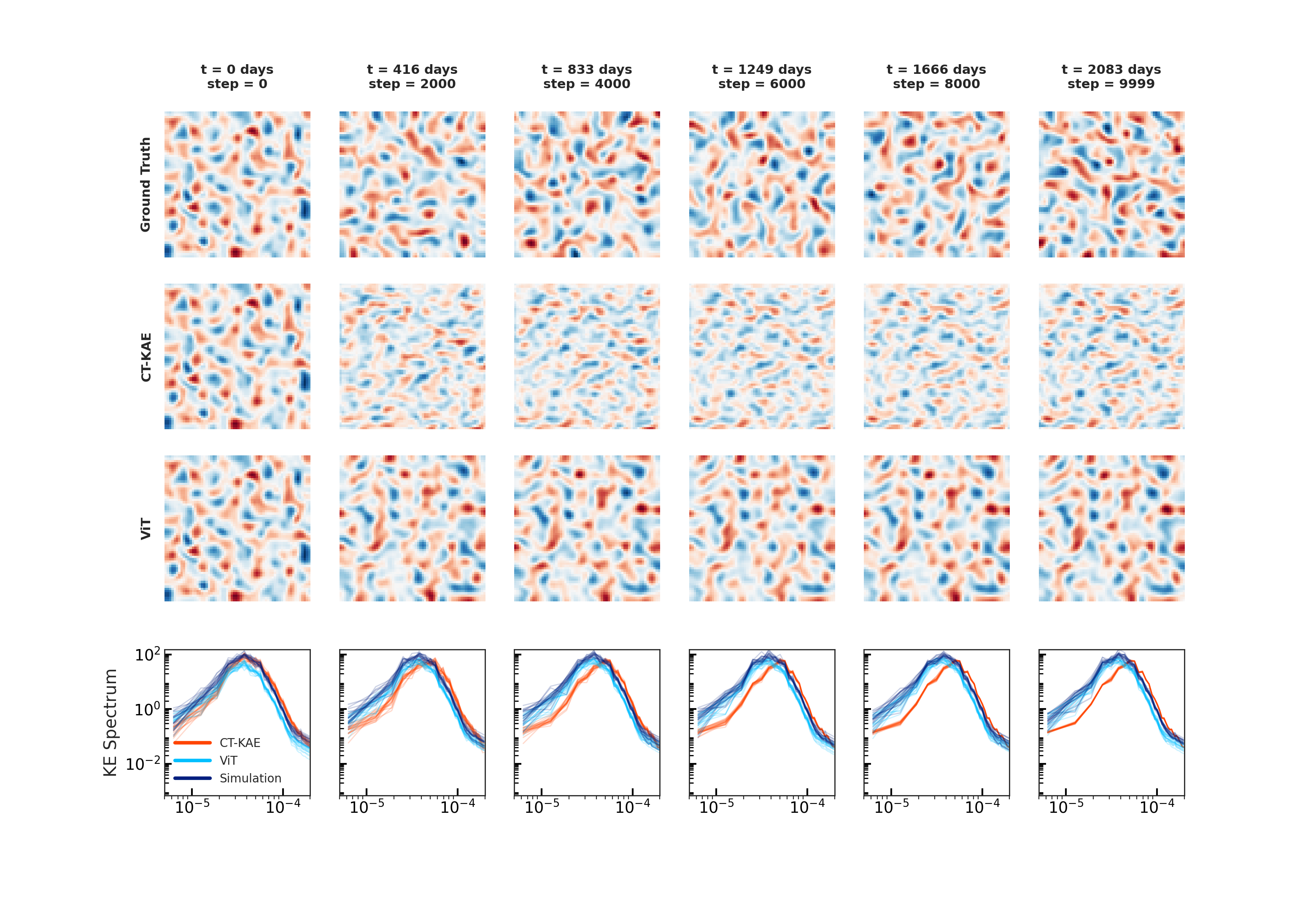}
    \caption{
    \textbf{Long-horizon qualitative comparison (2083 days).}
    \textbf{Top row:} Ground truth QG potential vorticity fields.
    \textbf{Middle row:} CT-KAE predictions rolled out for 9999 steps ($\Delta t = 5$h) without reinitialization.
    \textbf{Bottom row:} Kinetic Energy (KE) spectra averaged over the rollout for ground truth (blue) and CT-KAE (orange).
    The KE spectrum demonstrates preservation of bulk energy at low wavenumbers, while high-wavenumber modes are gradually attenuated. This controlled dissipation contributes to long-horizon stability, though it also indicates partial loss of fine-scale turbulent structures. Improving the capture of high-frequency dynamics while maintaining stability is a promising direction for future work.}
    \label{fig:overall_predicitons}
\end{figure*}

\begin{figure*}[t]
    \centering
    \begin{subfigure}[b]{0.32\linewidth}
        \includegraphics[width=\linewidth]{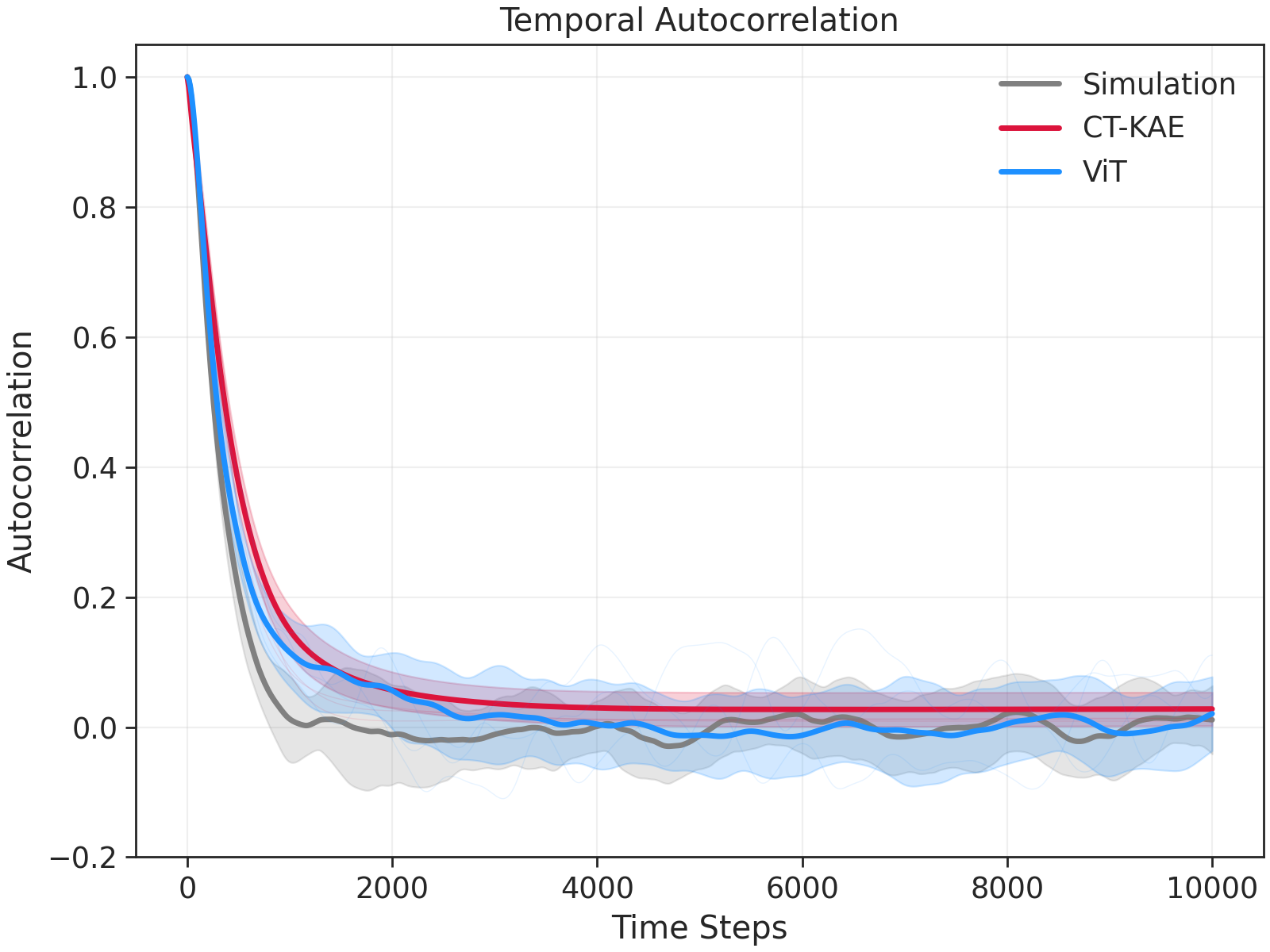}
        \caption{Temporal autocorrelation.}
    \end{subfigure}
    \hfill
    \begin{subfigure}[b]{0.32\linewidth}
        \includegraphics[width=\linewidth]{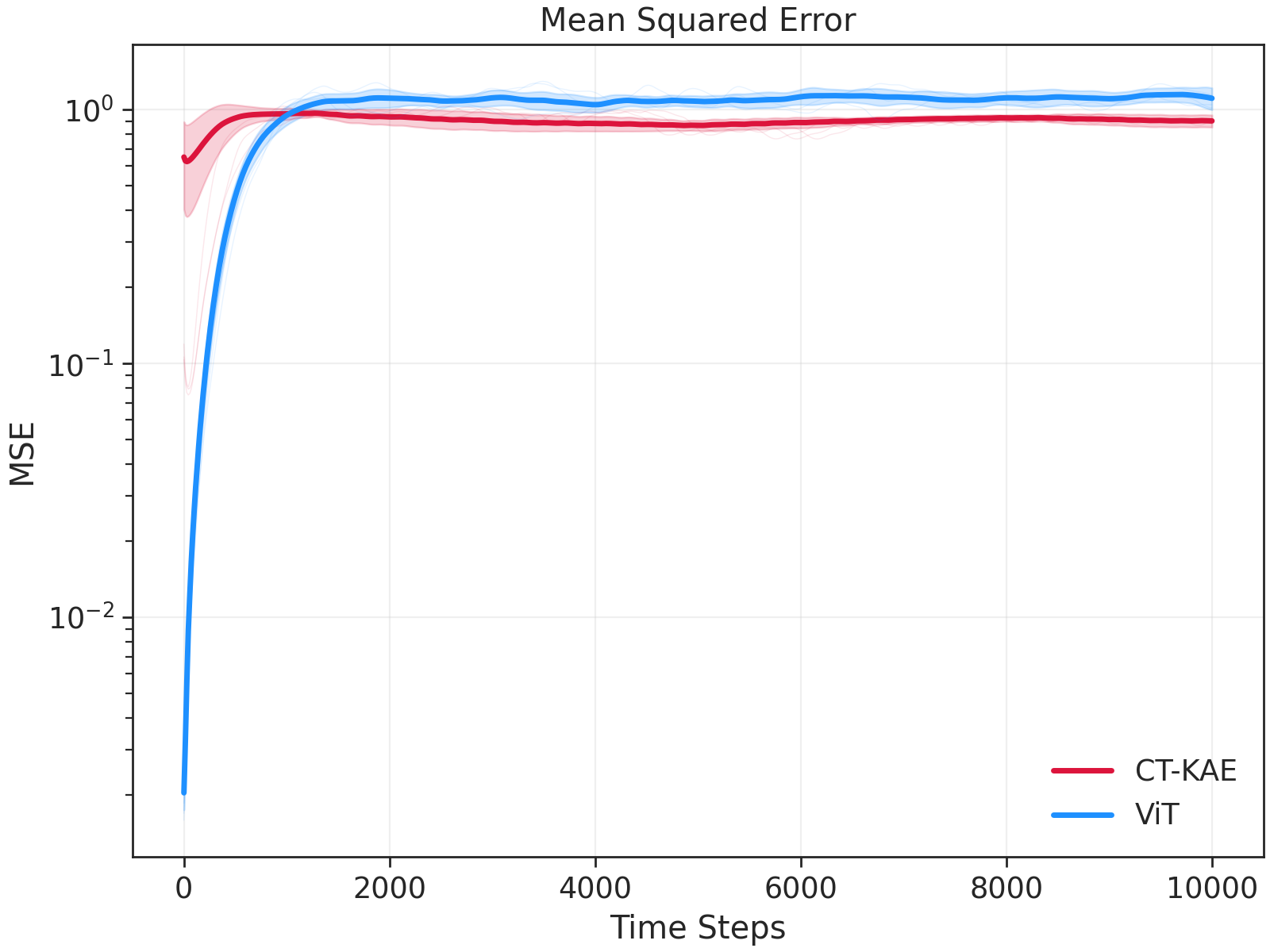}
        \caption{MSE over rollout.}
    \end{subfigure}
    \hfill
    \begin{subfigure}[b]{0.32\linewidth}
        \includegraphics[width=\linewidth]{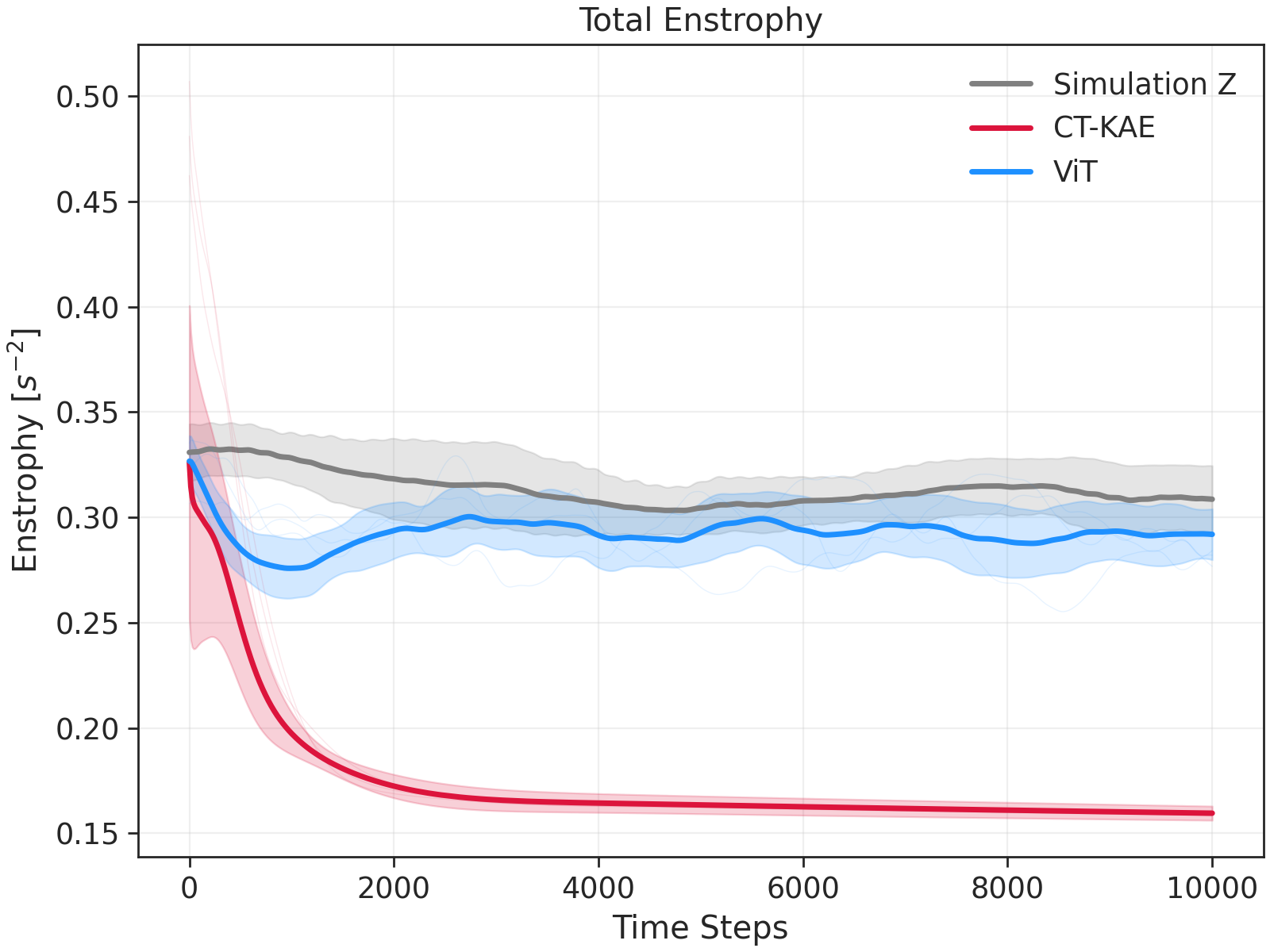}
        \caption{Enstrophy evolution.}
    \end{subfigure}
    \caption{
    \textbf{Stability over long horizons (20 trajectories).}
    (a) CT-KAE preserves large-scale persistence patterns.
    (b) MSE remains bounded over 10,000 rollout steps.
    (c) Enstrophy evolves without runaway amplification.}
    \label{fig:metrics}
\end{figure*}

Beyond the Lyapunov time, fidelity is assessed via preservation of the invariant measure, quantified through kinetic energy spectra, enstrophy evolution, and temporal autocorrelation. Figure~\ref{fig:overall_predicitons} shows that CT-KAE maintains large-scale vortical structures while mildly attenuating high-frequency modes. The KE spectrum confirms preservation of bulk energy at low wavenumbers without artificial amplification. Figure~\ref{fig:metrics}(b) shows bounded MSE growth over 10,000 steps, and enstrophy remains controlled.

We quantify energy stability using normalized drift,
\begin{equation}
\text{Drift} = \frac{E_T - E_0}{E_0},
\end{equation}
CT-KAE exhibits controlled negative drift ($-49.4\%$ KE, $-34.6\%$ enstrophy), consistent with mild dissipation, whereas ViT shows positive drift indicative of unphysical energy growth. Notably, the autoregressive ViT remains numerically stable over 10,000 steps and achieves competitive long-horizon RMSE, indicating that large Transformer models can learn partially self-regularizing dynamics. However, its positive error growth rate and energy drift suggest less controlled asymptotic behavior compared to CT-KAE.

\subsection{Operational Advantages of Continuous-Time Dynamics}




The matrix exponential formulation, in Equation \ref{eq:matrix_exp}
provides two practical advantages over discrete autoregressive models.
\paragraph{Efficiency.}
Inference reduces to a single latent matrix–vector multiplication, yielding sub-millisecond per-step prediction on an NVIDIA RTX4090 GPU, approximately $300\times$ faster than the pseudo-spectral solver. Computational cost scales with latent dimension $d$, not spatial resolution, enabling efficient ensemble forecasting.
\paragraph{Temporal Resolution Invariance.}
Training is performed at $\Delta t = 5$h and evaluated at 1h and 10h without retraining. RMSE remains stable across these unseen discretizations (Appendix~\ref{app:delta_t}), confirming resolution invariance inherent to the continuous-time formulation.

\begin{table}[ht]
\centering
\centering
\caption{Long-horizon performance at 2083-day horizon (10,000 steps). Averaged across trajectories.}
\label{tab:metrics}
\begin{tabular}{lcc}
\toprule
\textbf{Metric} & \textbf{CT-KAE} & \textbf{ViT (AR)} \\
\midrule
RMSE $\downarrow$ & \textbf{0.938 $\pm$ 0.031} & 0.993 $\pm$ 0.033 \\
ACC $\uparrow$ & \textbf{0.033} & 0.004 \\
KE Drift (\%) $\to 0$ & -49.4 & 23.3 \\
Enstrophy Drift (\%) $\to 0$ & \textbf{-34.6} & 37.0 \\
Error Growth Rate $\downarrow$ & \textbf{-0.0267} & 0.0217 \\
\bottomrule
\end{tabular}
\end{table}
\vspace{-0.5em}
\section{Conclusion}


We show that structured continuous-time latent dynamics enable stable long-horizon emulation of chaotic ocean flows at negligible computational cost. By enforcing linear operator structure in latent space, CT-KAE achieves bounded error growth and preservation of bulk physical statistics over multi-year rollouts. Rather than optimizing short-term trajectory alignment, the model approximates the invariant statistical structure of the system. The resulting combination of stability, efficiency, and temporal resolution invariance makes continuous-time Koopman surrogates a promising foundation for scalable physical–machine learning climate models.





\bibliography{iclr2026_conference}
\bibliographystyle{iclr2026_conference}

\appendix
\section{Appendix}

\subsection{Model Architecture}
\label{app:arch}

The Continuous-Time Koopman Autoencoder (CT-KAE) is designed to map high-dimensional turbulent states into a low-dimensional manifold where the dynamics can be approximated by a linear Ordinary Differential Equation (ODE). The architecture consists of three primary components: a history-aware encoder, a continuous-time latent evolution module, and a generative decoder.

\textbf{Encoder.} The encoder $\mathcal{E}$ projects the physical state $x_t \in \mathbb{R}^{H \times W \times C}$ into a latent vector $z_t \in \mathbb{R}^{d}$. To capture the temporal derivatives necessary for defining the flow, we utilize a dual-input strategy. The encoder processes both the current state $x_t$ and the history $x_{t-1}$.
The backbone consists of Pre-Activation Residual Blocks equipped with Spectral Normalization to stabilize the Lipschitz constant of the mapping. To allow the model to focus on active vortex shedding regions while ignoring background noise, we incorporate a Convolutional Block Attention Module (CBAM) at the bottleneck.
The final latent state is obtained via a linear projection:
\begin{equation}
    z_t = \mathcal{E}(x_t, x_{t-1})
\end{equation}

\textbf{Decoder.} The decoder $\mathcal{D}$ mirrors the encoder structure, utilizing pixel-shuffle upsampling blocks to reconstruct the physical field $\hat{x}_t = \mathcal{D}(z_t)$. To ensure spatial consistency and mitigate spectral bias, we employ \textit{Coordinate Injection}, concatenating a fixed grid of $(x, y)$ coordinates to the feature maps before the final reconstruction layer.

\subsection{Continuous-Time Latent Dynamics}
\label{app:dynamics}

Unlike discrete-time recurrent models, the CT-KAE governs the latent space via a learned linear ODE. This allows for resolution-invariant forecasting and theoretical stability analysis.

\textbf{Operator Decomposition.} The evolution of the latent state $z$ is defined as $\frac{dz}{dt} = \mathbf{K}z$. To ensure stable learning of the Koopman operator $\mathbf{K} \in \mathbb{R}^{d \times d}$, we decompose it into skew-symmetric and symmetric components:
\begin{equation}
    \mathbf{K} = \mathbf{K}_{\text{skew}} + \mathbf{K}_{\text{sym}}
\end{equation}
where $\mathbf{K}_{\text{skew}} = \frac{1}{2}(\mathbf{W} - \mathbf{W}^T)$ and $\mathbf{K}_{\text{sym}} = \frac{1}{2}(\mathbf{D} + \mathbf{D}^T)$.
\begin{itemize}
    \item $\mathbf{K}_{\text{skew}}$ corresponds to purely imaginary eigenvalues, capturing the oscillatory/advective dynamics (vortex rotation) while preserving energy.
    \item $\mathbf{K}_{\text{sym}}$ corresponds to real eigenvalues, capturing the growth and decay of modes (dissipation).
\end{itemize}
This decomposition prevents the "exploding gradient" problem common in RNNs by allowing explicit control over the system's energy.

\textbf{Integration.} During training, the trajectory is integrated using a 4th-order Runge-Kutta (RK4) scheme for numerical stability over short horizons. For inference and long-term forecasting, we utilize the matrix exponential solution:
\begin{equation}
    z(t + \tau) = \exp(\mathbf{K}\tau) z(t)
\end{equation}
This formulation allows the model to predict the ocean state at arbitrary continuous time-steps $\tau$, independent of the training data's temporal resolution.

\subsection{Physics-Informed Training Objective}
\label{app:loss}

To enforce physical consistency and prevent the model from learning trivial identity mappings, we minimize a composite loss function $\mathcal{L}_{total}$:

\begin{equation}
    \mathcal{L}_{total} = \mathcal{L}_{recon} + \alpha \mathcal{L}_{pred} + \beta \mathcal{L}_{latent} + \gamma \mathcal{L}_{phys}
\end{equation}

\textbf{1. Reconstruction \& Prediction ($\mathcal{L}_{recon}, \mathcal{L}_{pred}$):} We calculate the Mean Squared Error (MSE) for the immediate reconstruction and the future states rolled out via the latent dynamics. We apply a gradient-weighted mask to penalize errors in high-gradient regions (e.g., shock fronts or eddy edges) more heavily than the quiescent background.

\textbf{2. Latent Regularization ($\mathcal{L}_{latent}$):} To prevent mode collapse—where the model utilizes only a fraction of the latent dimensions—we enforce a covariance whitening loss. This penalizes off-diagonal correlations in the latent batch covariance matrix, encouraging an orthogonal and information-rich latent space. Additionally, we apply an eigenvalue repulsion term to disperse the learned frequencies of $\mathbf{K}$.

\textbf{3. Physics Consistency ($\mathcal{L}_{phys}$):} We incorporate domain-specific constraints:
\begin{itemize}
    \item \textit{Sobolev Loss:} Penalizes the error in the spatial gradients ($\nabla \hat{x}$ vs $\nabla x$) to ensure the vorticity fields remain sharp.
    \item \textit{Spectral Loss:} Minimizes the error in the Fourier domain (FFT) to ensure the energy cascade of the turbulent flow is preserved across wavenumbers.
\end{itemize}

\subsection{Hyperparameters}
\label{app:hyperparams}

\begin{table}[h]
\centering
\begin{tabular}{ll}
\hline
\textbf{Parameter} & \textbf{Value} \\
\hline
Latent Dimension ($d$) & 128 \\
Batch Size & 32 \\
Optimizer & AdamW \\
Learning Rate & $2 \times 10^{-4}$ \\
Weight Decay & $1 \times 10^{-5}$ \\
ODE Integration & Runge-Kutta 4 (Train), Matrix Exp (Test) \\
Activation Function & SiLU (Swish) \\
Spectral Normalization & Applied to all Conv/Linear layers \\
\hline
\end{tabular}
\caption{Hyperparameters used for the CT-KAE experiments.}
\label{tab:hyperparams}
\end{table}

\subsection{Temporal Invariance}
\label{app:delta_t}
\begin{figure*}
    \centering
    \includegraphics[width=\linewidth]{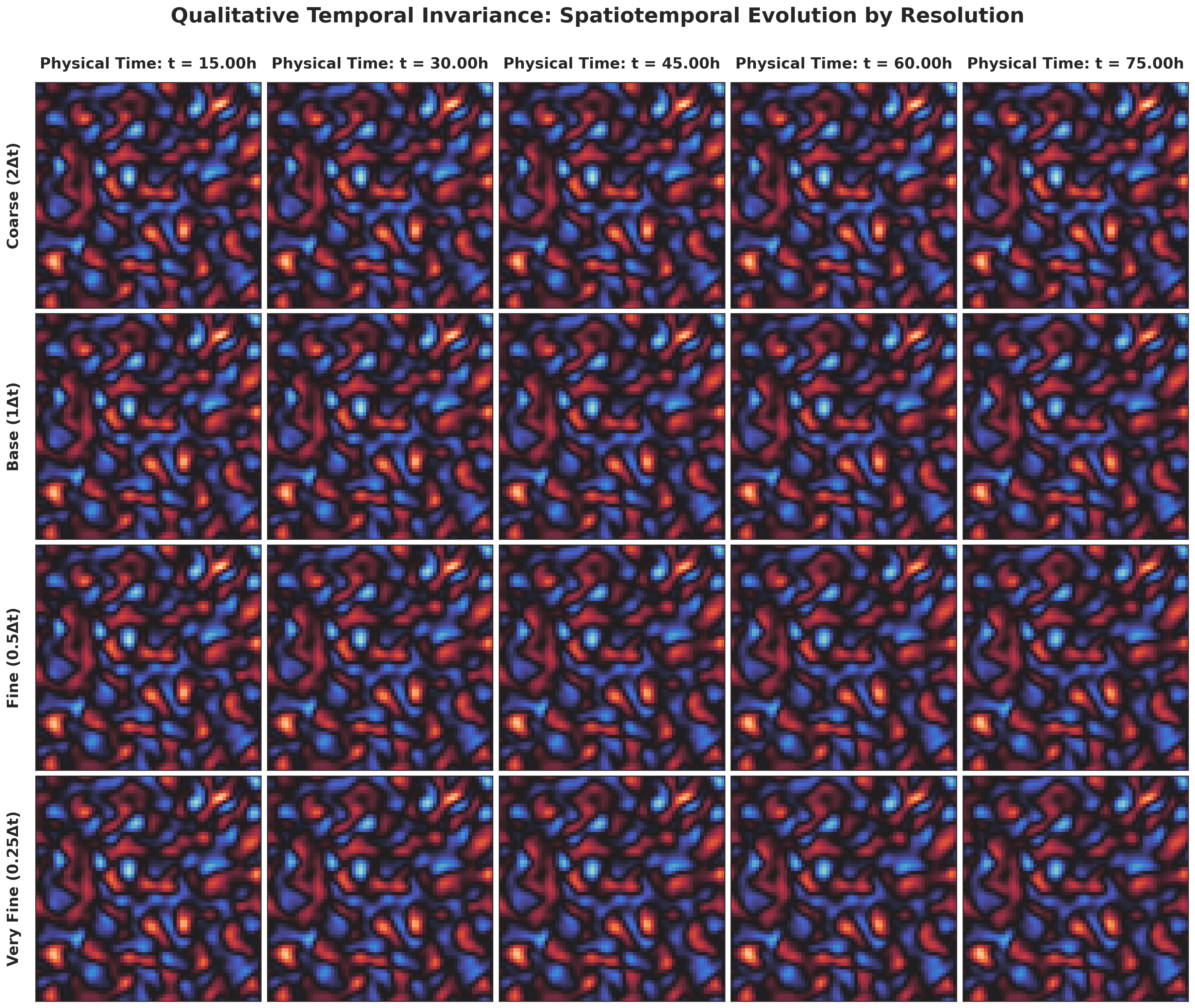}
    \caption{\textbf{Temporal resolution invariance.} CT-KAE trained at $\Delta t=5$h is evaluated at different integration resolutions ($1h$ and $10h$) using the continuous-time formulation. Each row corresponds to a different query time-step. The qualitative similarity across resolutions demonstrates that latent dynamics generalize across temporal discretizations without retraining, validating the matrix exponential formulation.}
    \label{fig:temporal_invariance}
\end{figure*}

To validate resolution invariance, we evaluate the trained model at unseen temporal discretizations (Figure \ref{fig:temporal_invariance}).

\subsection{Learned Operator Spectrum}

To analyze the stability properties of the learned Koopman operator $\mathbf{K}$, we compute its eigenvalue spectrum after training. The majority of eigenvalues lie in the left half of the complex plane, indicating dissipative behavior, while a small subset lies near the imaginary axis, corresponding to oscillatory modes that preserve large-scale rotational structures.

\begin{figure*}
    \centering
    \includegraphics[width=0.6\linewidth]{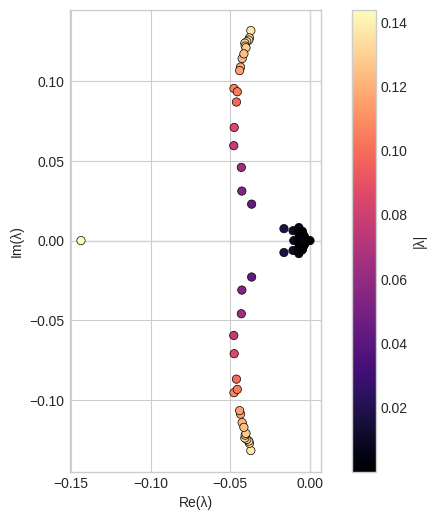}
    \caption{\textbf{Eigenvalue spectrum of the learned Koopman operator $\mathbf{K}$.} Eigenvalues are plotted in the complex plane. The majority lie in the left half-plane (negative real part), indicating dissipative modes, while a subset lies close to the imaginary axis, corresponding to oscillatory dynamics. The learned operator spectrum exhibits no eigenvalues with positive real part, which is consistent with the empirically observed bounded latent trajectories. While this does not constitute a formal stability guarantee of the decoded dynamics, it provides structural evidence supporting long-horizon stability as from Section \ref{section:results}.}
    \label{fig:operator_spectrum}
    
\end{figure*}

This spectral structure explains the bounded long-horizon error growth observed in Section~\ref{section:results}. Unlike autoregressive nonlinear predictors, where instability emerges implicitly, the continuous-time formulation provides explicit control over mode growth and decay.
To explicitly analyze stability, we inspect the learned operator spectrum (Figure \ref{fig:operator_spectrum}).



\end{document}

%% file: math_commands.tex
\usepackage{amsmath,amsfonts,bm}

\def\eqref#1{equation~\ref{#1}}

\def\1{\bm{1}}

\DeclareMathAlphabet{\mathsfit}{\encodingdefault}{\sfdefault}{m}{sl}
\SetMathAlphabet{\mathsfit}{bold}{\encodingdefault}{\sfdefault}{bx}{n}



%% file: iclr2026_conference.bib
@article{lee2019deep,
  author = {Sangseung Lee and Donghyun You},
  title = {Data-driven prediction of unsteady flow over a circular cylinder using deep learning},
  journal = {Journal of Fluid Mechanics},
  year = {2019},
  volume = {879},
  pages = {217-254}
}

@article{deo2023cnn,
  author = {Indu Kant Deo and Rui Gao and Rajeev Jaiman},
  title = {Combined space–time reduced-order model with three-dimensional deep convolution for extrapolating fluid dynamics},
  journal = {Physics of Fluids},
  year = {2023},
  volume = {35},
  issue = {4}
}

@inproceedings{chattopadhyay2020deep,
  title={Deep spatial transformers for autoregressive data-driven forecasting of geophysical turbulence},
  author={Chattopadhyay, Ashesh and Mustafa, Mustafa and Hassanzadeh, Pedram and Kashinath, Karthik},
  booktitle={Proceedings of the 10th international conference on climate informatics},
  pages={106--112},
  year={2020}
}

@article{pedersen2025thermalizer,
  title={Thermalizer: Stable autoregressive neural emulation of spatiotemporal chaos},
  author={Pedersen, Chris and Zanna, Laure and Bruna, Joan},
  journal={arXiv preprint arXiv:2503.18731},
  year={2025}
}

@article{besabe2025data,
  title={Data-driven reduced order modeling of a two-layer quasi-geostrophic ocean model},
  author={Besabe, Lander and Girfoglio, Michele and Quaini, Annalisa and Rozza, Gianluigi},
  journal={Results in Engineering},
  volume={25},
  pages={103691},
  year={2025},
  publisher={Elsevier}
}

@misc{grozavescu2026continuous,
      title={Koopman Autoencoders with Continuous-Time Latent Dynamics for Fluid Dynamics Forecasting}, 
      author={Rares Grozavescu and Pengyu Zhang and Mark Girolami and Etienne Meunier },
      year={2026},
      eprint={2602.02832},
      archivePrefix={arXiv},
      primaryClass={cs.LG} 
}

@article{rajabi2025sea,
  title={Sea level forecasting using deep recurrent neural networks with high-resolution hydrodynamic model},
  author={Rajabi-Kiasari, Saeed and Ellmann, Artu and Delpeche-Ellmann, Nicole},
  journal={Applied Ocean Research},
  volume={157},
  pages={104496},
  year={2025},
  publisher={Elsevier}
}

@article{madec2015nemo,
  title={NEMO ocean engine},
  author={Madec, Gurvan and others},
  year={2015},
  publisher={Institut Pierre-Simon Laplace}
}

@article{marshall1997finite,
  title={A finite-volume, incompressible Navier Stokes model for studies of the ocean on parallel computers},
  author={Marshall, John and Adcroft, Alistair and Hill, Chris and Perelman, Lev and Heisey, Curt},
  journal={Journal of Geophysical Research: Oceans},
  volume={102},
  number={C3},
  pages={5753--5766},
  year={1997},
  publisher={Wiley Online Library}
}

@article{bi2023accurate,
  title={Accurate medium-range global weather forecasting with 3D neural networks},
  author={Bi, Kaifeng and Xie, Lingxi and Zhang, Hengheng and Chen, Xin and Gu, Xiaotao and Tian, Qi},
  journal={Nature},
  volume={619},
  number={7970},
  pages={533--538},
  year={2023},
  publisher={Nature Publishing Group UK London}
}

@article{lam2023learning,
  title={Learning skillful medium-range global weather forecasting},
  author={Lam, Remi and Sanchez-Gonzalez, Alvaro and Willson, Matthew and Wirnsberger, Peter and Fortunato, Meire and Alet, Ferran and Ravuri, Suman and Ewalds, Timo and Eaton-Rosen, Zach and Hu, Weihua and others},
  journal={Science},
  volume={382},
  number={6677},
  pages={1416--1421},
  year={2023},
  publisher={American Association for the Advancement of Science}
}

@article{pathak2022fourcastnet,
  title={Fourcastnet: A global data-driven high-resolution weather model using adaptive fourier neural operators},
  author={Pathak, Jaideep and Subramanian, Shashank and Harrington, Peter and Raja, Sanjeev and Chattopadhyay, Ashesh and Mardani, Morteza and Kurth, Thorsten and Hall, David and Li, Zongyi and Azizzadenesheli, Kamyar and others},
  journal={arXiv preprint arXiv:2202.11214},
  year={2022}
}

@article{ross2023benchmarking,
  title={Benchmarking of machine learning ocean subgrid parameterizations in an idealized model},
  author={Ross, Andrew and Li, Ziwei and Perezhogin, Pavel and Fernandez-Granda, Carlos and Zanna, Laure},
  journal={Journal of Advances in Modeling Earth Systems},
  volume={15},
  number={1},
  year={2023},
  publisher={Wiley Online Library}
}

@article{frezat2022posteriori,
  title={A posteriori learning for quasi-geostrophic turbulence parametrization},
  author={Frezat, Hugo and Le Sommer, Julien and Fablet, Ronan and Balarac, Guillaume and Lguensat, Redouane},
  journal={Journal of Advances in Modeling Earth Systems},
  volume={14},
  number={11},
  pages={e2022MS003124},
  year={2022},
  publisher={Wiley Online Library}
}

@book{majda2006nonlinear,
  title={Nonlinear dynamics and statistical theories for basic geophysical flows},
  author={Majda, Andrew and Wang, Xiaoming},
  year={2006},
  publisher={Cambridge University Press}
}

@article{medjo2000numerical,
  title={Numerical simulations of a two-layer quasi-geostrophic equation of the ocean},
  author={Medjo, T Tachim},
  journal={SIAM journal on numerical analysis},
  volume={37},
  number={6},
  pages={2005--2022},
  year={2000},
  publisher={SIAM}
}

@article{koopman1931,
  author = {Koopman, B. O.},
  title = {Hamiltonian systems and transformation in hilbert space},
  journal = {Proceedings of the National Academy of Sciences},
  year = {1931},
  volume = {17},
  issue = {5},
  pages = {315-318}
}

@article{hemmasian2023transformer,
  author = {AmirPouya Hemmasian and Amir Barati Farimani},
  title = {Reduced-order modeling of fluid flows with transformers},
  journal = {Physics of Fluids},
  year = {2023},
  volume = {35},
  issue = {5}
}

@article{lusch2018koopman,
  author = {Bethany Lusch and J. Nathan Kutz and Steven L. Brunton},
  title = {Deep learning for universal linear embeddings of nonlinear dynamics},
  journal = {Nature Communications},
  year = {2018},
  volume = {9},
  issue = {1}
}

@article{bolton2019applications,
  title={Applications of deep learning to ocean data inference and subgrid parameterization},
  author={Bolton, Thomas and Zanna, Laure},
  journal={Journal of Advances in Modeling Earth Systems},
  volume={11},
  number={1},
  pages={376--399},
  year={2019},
  publisher={Wiley Online Library}
}
